\documentclass[runningheads,a4paper]{llncs}

\usepackage{amssymb}
\usepackage{graphicx}

\usepackage{amsbsy}
\usepackage{subfigure}

\usepackage{url}
\newcommand{\keywords}[1]{\par\addvspace\baselineskip
\noindent\keywordname\enspace\ignorespaces#1}

\begin{document}

\mainmatter  

\title{Weighted Data Normalization Based on Eigenvalues for Artificial Neural Network Classification}

\titlerunning{Weighted Data Normalization Based on Eigenvalues}

%
%
\author{Qingjiu Zhang, Shiliang Sun}
\authorrunning{Qingjiu Zhang, Shiliang Sun}

\institute{Department of Computer Science and Technology, East China
Normal University, \\500 Dongchuan Road, Shanghai 200241, P.R.
China\\qjzh08@gmail.com, slsun@cs.ecnu.edu.cn}

%
%

\toctitle{Lecture Notes in Computer Science} \tocauthor{Authors'
Instructions} \maketitle

\begin{abstract}
Artificial neural network (ANN) is a very useful tool in solving
learning problems. Boosting the performances of ANN can  be mainly
concluded from two aspects: optimizing the architecture of ANN and
normalizing the raw data for ANN. In this paper, a novel method
which improves the effects of ANN by preprocessing the raw data is
proposed. It totally leverages the fact that different features
should play different roles. The raw data set is firstly
preprocessed by principle component analysis (PCA), and then its
principle components are weighted by their corresponding
eigenvalues. Several aspects of analysis are carried out to analyze
its theory  and the applicable occasions.  Three classification
problems are launched by an active learning algorithm to verify the
proposed method. From the empirical results, conclusion comes to the
fact that the proposed method can significantly improve the
performance of ANN.

\end{abstract}

\section{Introduction}
Artificial neural network (ANN), usually called ``neural network''
(NN), is an information processing paradigm inspired by biological
nervous systems. The mathematical model of ANN is constructed by
lots of ``neurons'' which often work together, usually in the form
of hierarchy, to solve learning problems. Each neuron has a
threshold function which can be continuous or discrete. ANN often
has several layers, the former layer's outputs are weighted and used
as the inputs of the next layer. The function of the latter layer
maps the inputs to its outputs which will be weighted again and used
as the inputs of the next layer. Therefore, when the threshold
functions are selected, all the efforts are to find the weights.

ANN has wide applications, and it also has its shortcomings. Almost all the learning problems, such as classification, progression and multitask learning~\cite{multitask} can be solved by ANN. Relative researches have show that ANN can
solve not only linear problems, but also nonlinear problems. Moreover, it has
been proven that ANN with three layers can fit any non-linear
problems~\cite{duda}. When an neuron of the neural network
fails, others can function without any problem by their parallel
characteristics. Although ANN has lots of advantages, it has some
defects. Many kinds of ANNs are prone to step into local minimum
problems, which means a single ANN is not a stable learner. Ensemble
techniques sometimes can overcome this kind of problem by combining
several ANNs~\cite{bagging,adaboost,ensemble1}. In addition, the settings of
ANN's parameters, such as the number of hidden neurons and the
learning rate, are strongly depended on the characters of the data
set. Therefore, there is none fixed rule to set them.  Usually,
expert knowledge plays an important role in setting those parameters
for concrete issues.

The improvements for neural network can be mainly divided into two
aspects: changing ANN's architecture and normalizing the original
data. Changing the architecture of neural network sometimes means
constructing new types of ANNs. Researches on ANN may start from the
single-layer neural network. Single-layer network has simple
input-output relations, thus sheds light on the research of
multi-layer network~\cite{nnpr}. Based on different theories and
thoughts, different types of ANNs have been put forward, such as
feedforward neural network, radial basis function (RBF) network,
hopfield network and so on. They are based on different mathematical
models which are suitable to solve different learning problems.

Normalizing the data is another manner to improve the accuracy of
ANN. In many neural network applications, raw data (not preprocessed
or not normalized) is used. However, raw data suffer lots of
problems including high dimensional and time-consuming problems. By
normalizing the data, ANN can get better effects and save much time
for training. Using correlation coefficients as weights for input
variables can significantly boost ANN~\cite{coefficient}. Song and
Kasabov also presented their preprocessing data method WDN-RBF for
radial basic function typed neural networks~\cite{wdn-rbf}.
Furlanello and Giuliani have normalized raw data by combining local
and global space transformations~\cite{combinepca}.

This paper focuses on the data normalization approach for ANN. A new
data normalization method WDNE (Weighted Data Normalization based on
Eigenvalues) which weights the data by eigenvalues is proposed. WDNE
is different from existing data normalization methods. It leverages
the fact that the features which have different potentials in
learning problems should play different roles. The data set is
weighed by the eigenvalues, which means some features are enhanced
while others are weakened. WDNE firstly uses principle component
analysis (PCA) to rebuild the data set to ensure the features are
uncorrelated. Then all the features are weighted by their
corresponding eigenvalues.

The rest of the paper is organized as follows. Section 2 describes
our proposed normalization process and gives the corresponding
analysis. Section 3 reports experimental results. At last,
conclusions are drawn in Section 4.

\section{WDNE and Analysis}
WDNE is based on PCA. It can be regarded as a method that induces
the weights in ANN to change, and it can also be regarded as an
approach which preprocesses the data before applied.
\subsection{PCA}
PCA is mathematically defined as an orthogonal linear transformation
that transforms a number of possibly correlated variables into a
smaller number of uncorrelated variables called principal
components. PCA can be used  to reduce the high dimensionality of
dimensional data and improve the predictive performance of some
machine learning methods~\cite{PCA1}. High dimensional data often
cause computational problems and run the risk of overfitting.
Furthermore, many redundant or highly correlated features may
probably cause a degradation of prediction accuracy. By simply
discarding some features which have little information, PCA can
eliminate the problem of high dimensionality.

The process of PCA  aims to transform a problem from its natural
space into another space, in which all the mapped features have
irrelevant relationships. Suppose that
\begin{math}\boldsymbol{X=\{x_1,x_2,...,x_n} \}\end{math} is the data set
which contains \begin{math}n\end{math} examples, and that the
problem has \begin{math}d\end{math} dimensions, which means every
example has
\begin{math}d\end{math} random variables. For example, a example can be expressed as \begin{math}\boldsymbol{x_i}=\{ x_i^1 ,x_i^2,...,x_i^d\}
\end{math}. When the data set \begin{math}\boldsymbol{X}\end{math} is available,
covariance matrix can be calculated in the form of
\begin{math}\boldsymbol{\sum=E\{(x-\mu)(x-\mu)}^T\}\end{math}, where \begin{math}E\end{math} means figuring
out the mathematical expectation, and
\begin{math}\boldsymbol{\mu}\end{math} is the mean of the examples.
\begin{math}\sum\end{math} is usually figured out by the way of
\begin{math}\sum=1/n\sum_{i=1}^{i=n}(\boldsymbol{x_i-\mu)(x_i-\mu)}^T\end{math}.
By mathematical calculating, the eigenvalues
\begin{math}\boldsymbol{\lambda}=\{\lambda_1,\lambda_2,...,\lambda_d\}\end{math} and the
eigenvectors
\begin{math}\boldsymbol{U=\{u_1,u_2,...,u_d\}}\end{math} of the
covariance matric \begin{math}\Sigma\end{math} can be obtained.
Subsequently, the initial example
\begin{math}\boldsymbol{x}\end{math} is mapped to a new space in the
form of
\begin{math}\boldsymbol{x'=U}^T\boldsymbol{x}\end{math}. In the new space, every
feature of the example is uncorrelated. Unless stated otherwise, the
\begin{math}k\end{math}th principle component will be taken to mean
the principle components with the \begin{math}k\end{math}th largest
eigenvalue~\cite{PCA}. By selectively choosing some principle
components, the high dimensional space is transformed into a fitting
subspace.
\subsection{ Weighted Data Normalization Based on Eigenvalues}
The basic idea of the proposed method is that the more useful a
component is the more important role it should play. The data are
weighted by the eigenvalue vector
\begin{math}\boldsymbol{\lambda}\end{math}. Every component \begin{math}x_i^j\end{math} of the input \begin{math}\boldsymbol{x_i}\end{math}
which is processed by PCA has a corresponding eigenvalue
\begin{math}\lambda_j\end{math}. \begin{math}\lambda_j\end{math}
has lots of potential contents, one of which is that it indicates
the variance of the component. The one which has the bigger
eigenvalue can clearly provide more information, and PCA  mainly
depends this point for dimensionality reduction. However, if the
data set is just processed by PCA,  all the selected components
still have equal roles in the training process. Actually, the
component providing more information should play more decisive  role
in solving learning problems. Therefore, all the components can be
weighted by their corresponding eigenvalues.

WDNE can be divided into two steps. Initially, it processes the data
by PCA. Then, the processed data are weighted by the eigenvalues
which are figured out in the first step. To avoid computational
problems, the eigenvalues are normalized before weighting the data.
The eigenvalues are normalized in the form of
\begin{math}\lambda_j'=\lambda_j/\lambda_1\end{math}, where
\begin{math}\lambda_1\end{math} is the biggest eigenvalue. Subsequently, each
feature is weighted by the corresponding eigenvalue in the form of
\begin{math}x_i^j=x_i^j\lambda_j'\end{math}. If the
original data set has high dimensionality, it can be reduced by
discarding some features which provide little information.
\subsection{Analysis}
WDNE  reinforces the principle components with large eigenvalues and
weakens the others. If the PCA-processed data are directly used, all
the components will play an equal role in the training process.
However, if the data are weighted by WDNE, the principle components
will play more important roles than the sub-principle components. It
can be apparently analyzed from the data processing. The principle
component multiplies a relative large constant, which means the
distance between elements will be enlarged. When applied into
learning problems, the principle components will play a more
important role in deciding the final hypotheses. Similarly, the
components which provide little information are  weakened, because
they can not provide good decisions. From this point of view, the
PCA, in which the selected components are weighted by one while the
unselected components are weighted by zero, can be seen as a special
situation of WDNE.

WDNE can also be regarded as a method which enhances the effects of
ANN by  weighting some of  the inputs.  Initially, the output
\begin{math}net=\boldsymbol{w}^T\boldsymbol{x}=\sum_{j=1}^dw_jx_j\end{math} of
the input layer is mapped to the function \begin{math}f\end{math} of
the hidden layer. If there is no WDNE, the output of the hidden
layer is
\begin{math}y=f(net)\end{math}. When WDNE is used, the output of the
hidden layer will be
\begin{math}y=f(net')=f(\boldsymbol{w}^T\boldsymbol{\Lambda}\boldsymbol{x})\end{math}, where \begin{math}\boldsymbol{\Lambda}\end{math} is a diagonal matrix with the eigenvalues as its values.
WDNE apparently does not change the architecture of ANN, instead, it
improves the performance by weighting the inputs. The process can be
explicitly described in  Fig. 1.
\begin{figure}
\centering
\includegraphics[height=5cm]{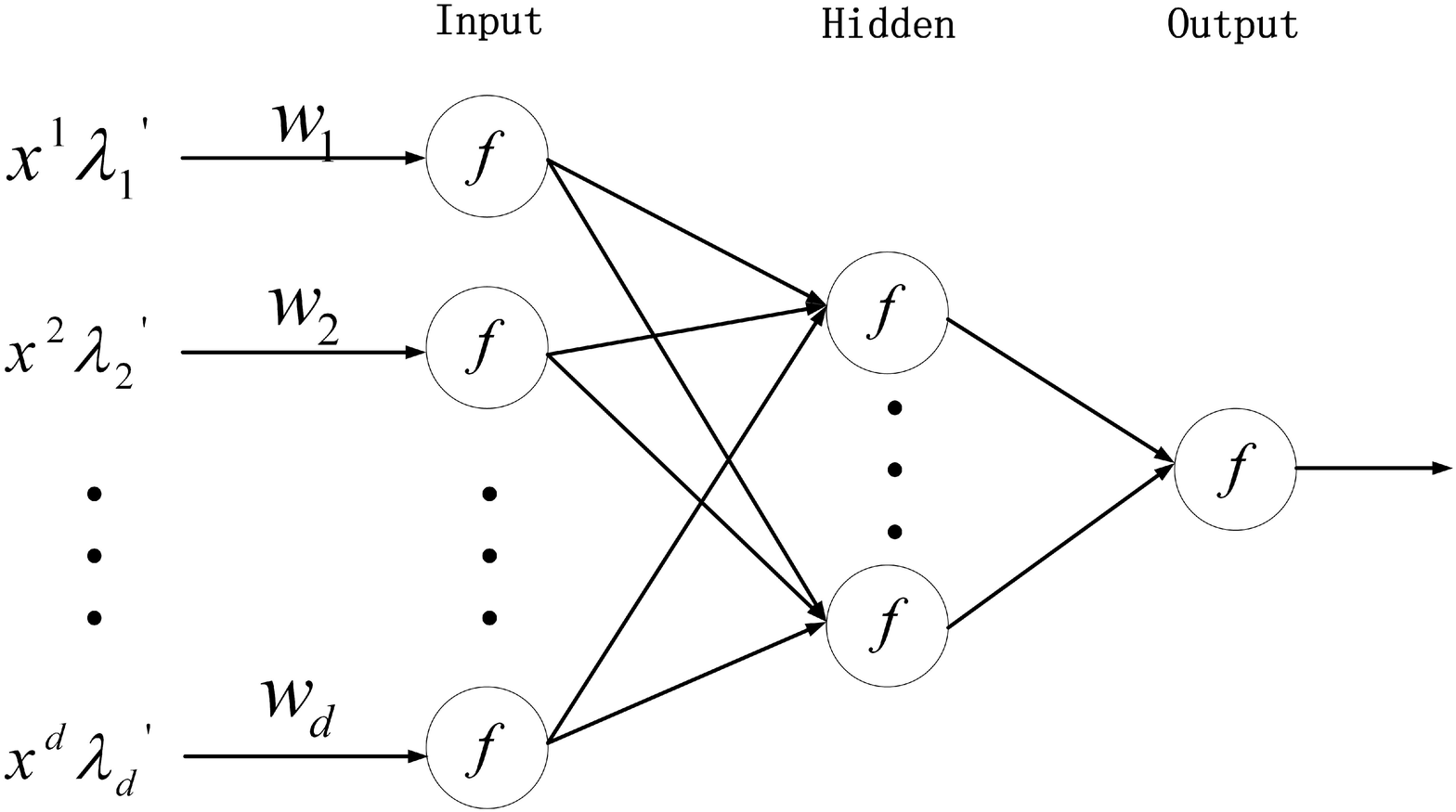}
\caption{WDNE.}
\end{figure}

WDNE is probably sensible to noise. Suppose some noisy features are
contained into the learning problem. The noisy features would be
transformed into a new space after PCA processing. If many of the
noisy ingredients are redistributed into the last several features,
the noisy effect is obviously reduced. However, if many of the noisy
ingredients are transformed into the first few features which
correspond to the biggest eigenvalues, the noisy effect will be
enhanced, and the performance of the recognition will be degraded.
Consequently, caution should be taken  before WDNE is applied to the
problem which may contain noises.
\section{Experiments}
Back propagation neural network is applied in the implemented
experiments. All of the data sets come from the UCI
repository\footnote{http://archive.ics.uci.edu/ml/}. The ANN has
three layers, and the number of hidden neurons is set as
\begin{equation}
    neuron\_number=\sqrt{n+m}+a\ ,
\end{equation}
where \begin{math}n\end{math} and \begin{math}m\end{math} denote the
number of input and output neurons of the networks respectively, and
\begin{math}a\end{math} is a constant ranging from one to ten.
In order to effectively boost WDNE, five ANNs are used in every
experiment, and the final hypothesis is the voted result by the five
committees.

Active learning algorithm and ten-fold cross validation (CV) are
lunched on the three experiments. Active learning aims to reduce the
number of labeled data for learning by selecting the most
informative examples~\cite{activelearning}. The active learning
algorithm used in this paper is for classification. Initially, only
a few labeled examples are prepared at hand. At every round of
iteration
\begin{math}\alpha\end{math} examples are selected from
\begin{math}\beta\end{math} candidates which are the most valuable ones
in the unlabeled examples pool. In order to more objectively
reflects the result, ten-fold CV is applied. The data set is divided
in to ten parts. One part is used as validation set, another part is
used as test data, and the others are used for training. The final
result is the average of the ten results.
\subsection{Comparative Methods}
In order to clearly evaluate the advantage of WDNE, three other
representations of data are implemented, and they are original data
(raw data), PCA processed data and negatively weighted data. The
descriptions of them are listed as follows:
\begin{itemize}
\item \textbf{Original data:} This data set is the original (raw) data set.
\item \textbf{PCA processed data:} This data set is just processed by PCA.
\item \textbf{Positively weighted data:} This data set is processed by WDNE. In other words, this data set is firstly processed by PCA, then all the components is weighted by the corresponding eigenvalues.
\item \textbf{Negatively weighted data:} This data set is firstly processed by PCA, then each component is weighted by the reversed eigenvalue \begin{math}\lambda_j'=\lambda_{d-j+1}/\lambda_1\end{math}.
\end{itemize}
All the compared data are trained by the back-propagation network
under the same setting of parameters.
\subsection{Heart Classification}
This is a two-class classification problem and the data set contains
150 positive examples and 120 negative examples. The initial
dimension space consists of 13 features. The labeled data used for
active learning algorithm initially contains six positive examples
and six negative examples. The size of the pool is 100. At every
round of iteration, four unlabeled examples are randomly selected
from 16 candidates which are regarded as the most valuable ones. The
final results can be seen from Fig. 2 (a).
\begin{figure}
\centering \subfigure[Heart classification]{
\includegraphics[width=165pt]{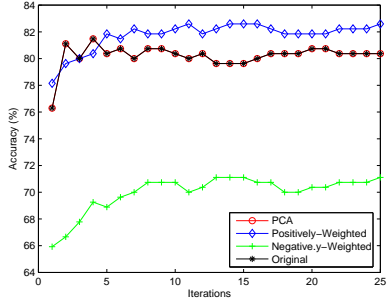}
\label{RSCurvefig:a}} \subfigure[Spam classification]{
\includegraphics[width=165pt]{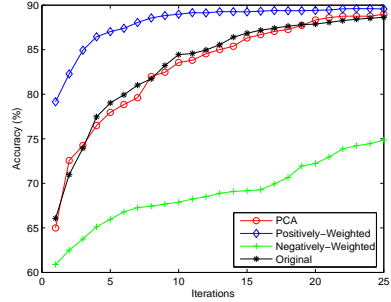}
\label{RSCurvefig:b}} \subfigure[Waveform classification]{
\includegraphics[width=165pt]{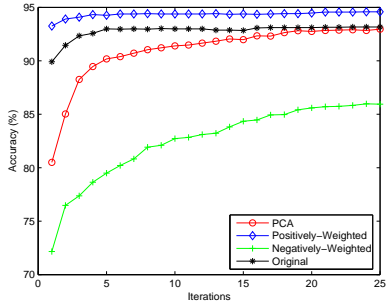}
\label{RSCurvefig:c}} \caption{Classification performances.}
\end{figure}

In Fig. 2 (a), the effect of the original data is almost equal to
that of the data which are just processed by PCA. Negatively
weighted data set has the worst performance, while positively
weighted data set is clearly the best representation.
\subsection{Spam Classification}
This data set is composed of 4501 examples which contains 1813
spams. The problem have totally 57 features without dimension
reduction. Initially, the active learning algorithm runs on 15
labeled examples which consists six positive ones and nine negative
ones. Before sampling, 32 unlabeled example which are seemed as the
most informative ones are chosen from the pool whose size is 240.
Subsequently, 10 examples are chosen from the 32 candidates at
random.


Fig. 2 (b) shows that WDNE is significantly superior to the other
three kinds of representations. As the results of last experiment,
the data set of negative weighted has the worst performance.
Although there is difference between the performances of original
data and PCA processed data, they almost have the same effect on the
whole.
\subsection{Waveform Classification}
This data set is also a binary classification which contains 1653
positive examples and 1655 negative examples. There are 40 features.
It is worth noting that 19 of the 40 features are noises. Six
positive examples and six negative examples are labeled at first. At
each round of iteration, the algorithm firstly chooses 32 most
informative examples from a pool whose size is 200. Subsequently,
eight examples are randomly selected from the 32 candidates.


Fig. 2 (c) shows the results. It is apparent to tell that the WDNE
performs the best and the negatively weighted data set plays the
worst. In this experiment, it is obviously that PCA processed data
are not as good as the original data. There may be several possible
causes of this phenomenon. (1) It is caused by the noises. (2) The
original representation of the data set is more suitable to solve
the learning problem in this experiment.
\section{Conclusions}
In this paper, a new data normalization approach WDNE which can be
used to improve the performances of neural networks is proposed.
WDNE does not optimize the architecture of the ANN. However, it
boosts the performance of ANN by preprocessing the data. PCA plays
an important role in this method. The components are weighted by the
corresponding eigenvalues of the covariance matrix. In order to
verify the proposed method, three other representations of data are
implemented in the experiments. All the utilized data sets come from
the UCI repository. The empirical results clearly show that WDNE is
an effective method for optimizing the performance of ANN.

Researches on WDNE requires further investigation. In this paper
WDNE is used for ANN, it may be applied to combine with other
approaches, such as distance metric learning (DML)~\cite{dml},
support vector machines (SVM)~\cite{svm}, etc. Moreover, WDNE may be
regraded as a way for feature selection.
\section*{Acknowledgments} This work is supported
in part by the National Natural Science Foundation of China under
Project $60703005$, and by Shanghai Educational Development
Foundation under Project $2007CG30$.

\end{document}